%% file: 0_main.tex
\pdfoutput=1

\documentclass[11pt]{article}

\usepackage[]{acl}

\usepackage{times}
\usepackage{latexsym}

\usepackage[T1]{fontenc}

\usepackage[utf8]{inputenc}

\usepackage{microtype}

\usepackage{algorithm}
\usepackage{algorithmic}
\usepackage{graphicx}
\usepackage{amsmath}
\usepackage{amssymb}
\usepackage{verbatim}
\usepackage{array}
\usepackage{booktabs} 
\usepackage{adjustbox}

\title{Disentangled Action Recognition with Knowledge Bases}

\author{Zhekun Luo \\
  \small{University of California, Berkeley} \\
  \small{zhekun\_luo@berkeley.edu} \\\And
  Shalini Ghosh\thanks{~~Work done prior to Amazon} \\
  \small{Amazon Alexa AI} \\
  \small{ghoshsha@amazon.com} \\ \And
  Devin Guillory \\
  \small{University of California, Berkeley} \\
  \small{Dguillory@berkeley.edu} \\\AND
  Keizo Kato\thanks{~~Work was done when Keizo Kato was at CMU}  \\
  \small{Fujitsu Laboratories Ltd.}\\
 \small{kato.keizo@jp.fujitsu.com} \\ \And
 Trevor Darrell  \\
  \small{University of California, Berkeley}\\
 \small{trevor@eecs.berkeley.edu} \And
 Huijuan Xu   \\
  \small{Pennsylvania State University}\\
 \small{hkx5063@psu.edu} }

\begin{document}
\maketitle
\begin{abstract}
     Action in video usually involves the interaction of human with objects. Action labels are typically composed of various combinations of verbs and nouns, but we may not have training data for all possible combinations. In this paper, we aim to improve the generalization ability of the compositional action recognition model to novel verbs or novel nouns that are unseen during training time, by leveraging the power of knowledge graphs. Previous work utilizes verb-noun compositional action nodes in the knowledge graph, making it inefficient to scale since the number of compositional action nodes grows quadratically with respect to the number of verbs and nouns. To address this issue, we propose our approach: Disentangled Action Recognition with Knowledge-bases (DARK), which leverages the inherent compositionality of actions. DARK trains a factorized model by first extracting disentangled feature representations for verbs and nouns, and then predicting classification weights using relations in external knowledge graphs. The type constraint between verb and noun is extracted from external knowledge bases and finally applied when composing actions. DARK has better scalability in the number of objects and verbs, and achieves state-of-the-art performance on the Charades dataset. We further propose a new benchmark split based on the Epic-kitchen dataset which is an order of magnitude bigger in the numbers of classes and samples, and benchmark various models on this benchmark.
     
\end{abstract}

\input{1_intro.tex}

\input{2_related.tex}

\input{3_method.tex}

\input{4_experiment.tex}

\input{5_conclusion.tex}

\clearpage
\newpage

\bibliography{anthology,custom}
\bibliographystyle{acl_natbib}

\clearpage

\appendix
\input{supp_1_implementation.tex}

\input{supp_2_epic.tex}

\input{supp_3_noun_KG.tex}

\input{supp_4_disentanglement.tex}

\end{document}

%% file: 1_intro.tex
\section{Introduction}
\label{sec:intro}

\begin{figure}[t]
\centering
\includegraphics[width=0.48\textwidth]{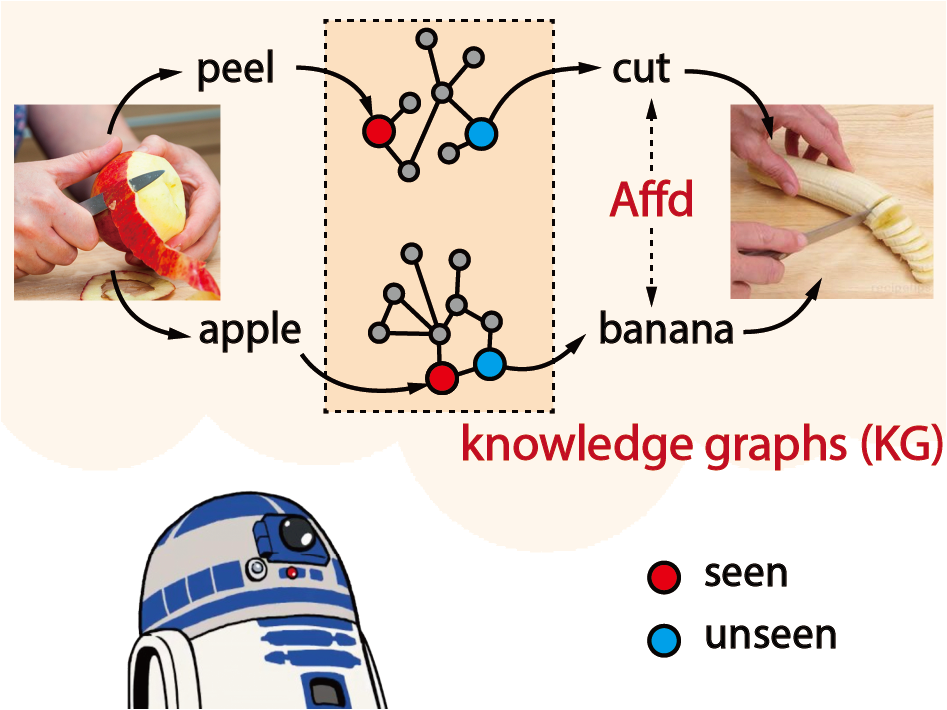}
\caption{DARK extracts disentangled features of verbs/nouns and leverages knowledge graphs (KG) to generate classifiers for unseen verbs and nouns. Predictions are then composed under constraints of object affordance priors (Affd) from knowledge bases (e.g. $banana~can~be~cut$). }

\label{fig:high level}
\end{figure} 

Understanding human-object interaction is crucial for modeling human behavior, and plays a key role in developing robotic agents that interact with humans. In videos, many of these interactions can be described using the combination of verbs and nouns, e.g, $move~chair$, $peel~apple$. Recently researchers have been focusing on the task of compositional action recognition with the goal of recognizing actions represented by such verb-noun pairs. The key challenge comes from the extremely large label space of various combinations. The number of possible verb-noun pairs grows quadratically with respect to the number of verbs and nouns. It is infeasible to collect training data for all possible actions. This motivates us to study the problem of zero-shot compositional action recognition, which aims to predict action with components beyond the vocabularies in train data.

To conduct zero-shot learning, we propose our Disentangled Action Recognition with Knowledge-bases (DARK), which leverages knowledge graphs. A knowledge graph (KG) encodes semantic relationships between verb or noun nodes. We apply graph convolutional network (GCN) on KG to predict classifier weights for unseen nodes in graphs.

Previous work~\cite{Kato_2018_ECCV} has explored using knowledge graphs for zero-shot compositional action learning. However, their model builds a graph containing verb, noun and compositional action nodes (which link verbs and nouns). It learns features of novel action nodes by propagating information from connected verb/noun nodes. The number of compositional action nodes during training is in the order of $\mathcal{O}(n^2)$, and the memory consumption may become prohibitively expensive as we scale this approach to large vocabularies. To overcome this issue, we propose to learn separate classifiers for verbs and nouns, which scales linearly with respect to the vocabulary size. 

Specifically, DARK extracts verb and noun features separately, and relies on separate verb and noun knowledge graphs to predict unseen concepts before composing the action label. Activity recognition is particularly well-suited for such a factorized approach, because nouns may be better captured using object detection-based approaches and verbs may be represented by motion. Compared to prior work \cite{Kato_2018_ECCV,cross_task} that use the same feature representation for both verb and noun, our separate features model noun and verb more precisely. In addition, we adopt disentanglement between the learned verb and noun features, so they compose more readily and improve generalization on unseen actions.

Though scalability is achieved using our factorized approach, verbs and nouns are actually not fully independent. For instance, the process of $sanding$ an object and $scrubbing$ an object are visually similar, however you are more likely to be $scrubbing$ a $car$ than $sanding$ it. Prior work~\cite{Kato_2018_ECCV} models the verb-noun relationships by constructing quadratic compositional action nodes. In our model, when composing action labels, we take object affordances~\cite{Affordance}, namely the commonsense relationship of verbs and nouns, into consideration. We extract affordance knowledge from a caption corpus and build a scoring component to consider the relationship between verbs and nouns, to further improve the generalization ability. The basic idea of our proposed model is illustrated in Figure~\ref{fig:high level}.

Furthermore, we investigate the evaluation of zero-shot compositional action recognition task and identify the drawback of existing metrics. With $N_v$ verbs, $N_n$ nouns, it constructs a $N_v \times N_n$ label space for possible actions. Among these actions, some are $invalid$ (e.g. $peel~a~car$) and some are $valid~but~not~presented$ in the dataset. In real-world applications, the model would need to make predictions in the whole $N_v \times N_n$ label space. But current evaluation protocols, implicitly or explicitly, only evaluate on compositional classes that are $valid~and~presented$ in the dataset, which does not reflect the real difficulty of this task. We propose a new setting, where predictions are made and evaluated in the full $N_v \times N_n$ label space. 

The Charades~\cite{charades} dataset is relatively small scale for testing zero-shot compositional action recognition \cite{Kato_2018_ECCV}. To promote further research, we propose a new benchmark based on the Epic-kitchen \cite{epic1,epic2} dataset, which is an order of magnitude bigger both in number of classes and sample size. The key contributions of our paper are:

\begin{enumerate}
\item We propose a novel factorized model that learns disentangled representation separately for verbs and nouns, facilitating scalability.

\item We further improve the model's generalization performance by learning the interaction constraints between verbs and nouns (affordance priors) from an external corpus.

\item We propose a new evaluation protocol for zero-shot compositional learnings, which better reflects the real-world application setting.

\item We propose a new large-scale benchmark based on the Epic-Kitchen dataset and achieve state-of-the-art results.
\end{enumerate}

%% file: 2_related.tex
\section{Related Work}
\label{sec:related}
  
{\bf Zero-shot learning with knowledge graphs:} Zero-shot learning has been widely studied in computer vision~\cite{AkataPHS16,LampertNH14,Lee_2018_CVPR,sahu2020cross,wang2019survey,XianLSA19}. We will focus on related work relevant to our approach. \cite{xiaolongwang} proposes to distill both the implicit knowledge representations (e.g., word embedding) and explicit relationships (e.g., knowledge graph) to learn a visual classifier for new classes through GCN \cite{GCN}. \cite{kampffmeyer2019rethinking} later proposes to augment the knowledge graph (KG) with dense connections which directly connects multi-hop relationship and distinguishes between parent and children nodes. The graph learning of our model mostly follows their work. Recently there have been improvements on GCN models. \cite{nayak2020zeroshot} designs a novel transformer GCN to learn representations based on common-sense KGs. \cite{Geng2019ExplainableZL} uses an attentive GCN together with an explanation generator to conduct explainable zero-shot learning. Instead of generating classifier for unseen classes directly, \cite{geng2020generative} uses a generative adversarial network to synthesize features for unseen classes to conduct classification. These directions could be potentially explored in our problem setting to further improve performance.~\cite{Gao_Zhang_Xu_2019} conducts zero-shot action recognition based on KGs, but unlike our problem setting, their verb-noun relationship is not compositional and objects are used as attributes to infer action. 

\noindent {\bf Compositional action recognition:} Many prior works aim to understand actions through interaction with objects. \cite{JunsongYan,HOI_knowledge} tackle zero-shot human-object interaction in images. \cite{cross_task} conduct weakly-supervised action recognition, leveraging compositionality of verb-noun pairs to decompose tasks into a set of verb/noun classifiers. This shares certain similarities with our factorized model, but it is not a zero-shot setting, nor does it enforce feature disentanglement. \cite{somethingelse} conduct zero-shot compositional action recognition, where individual verb/noun concepts have been seen during training but not in the same interaction with each other. Although it cannot deal with unseen verbs or nouns, using object detector to explicitly model object features inspires our approach. One of the closest works to our proposed approach is \cite{Kato_2018_ECCV}. It constructs a KG that contains verb nodes, noun nodes and compositional action nodes, and learns the feature representation for each action node to match visual features. Novel actions' features are inferred jointly during training through GCN. The number of action nodes grows quadratically with respect to the number of nouns or verbs, which makes this approach difficult to scale, especially considering that GCN's forward pass needs to learn all features simultaneously.

%% file: 3_method.tex
\section{Method}
\label{sec:method}

\begin{figure*}[hbtp]
\centering
\includegraphics[width=\textwidth]{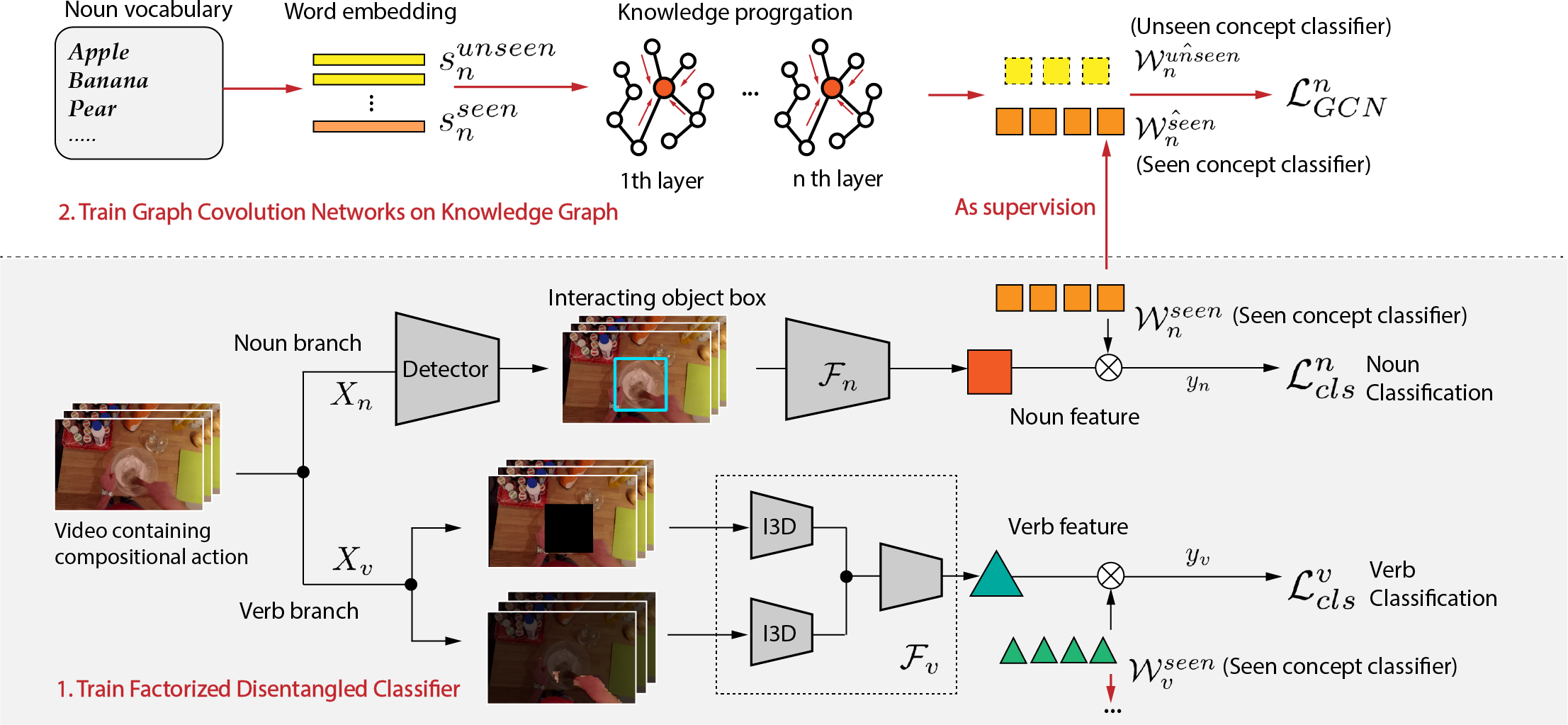}
\caption{Overall training process. We first jointly train factorized disentangled feature extractor $\mathcal{F}_v$, $\mathcal{F}_n$, and classifier weights for seen class $\mathcal{W}_n^{seen}$, $\mathcal{W}_v^{seen}$. We then take word embedding $s_{seen}, s_{unseen}$ as input, and use GCN to predict classifier weights of unseen class $\hat{\mathcal{W}_n^{unseen}}$, $\hat{\mathcal{W}_v^{unseen}}$ based on knowledge graphs (KGs). Note that the same GCN learning applies to both verbs~/~nouns, and in the figure we only show the noun's part for brevity. }

\label{fig:training}
\end{figure*}

We propose DARK -- Disentangled Action Recognition with Knowledge-bases (Figure~\ref{fig:training}). It extracts disentangled feature representations for verbs and nouns, then predicts classifier weights for unseen components using knowledge graphs, and composes them under object affordance priors. 

\subsection{Factorized verb-noun classifier}
\label{sec:method:factorized}

Given a video $X$, we first use a verb feature extractor $\mathcal{F}_v$ to extract verb feature, and a noun feature extractor $\mathcal{F}_n$ for noun feature. Subsequently, we learn one-layer predictors $\mathcal{W}_v^{seen}$ and $\mathcal{W}_n^{seen}$ for predicting the final verb/noun class. 
$\mathcal{F}_v$, $\mathcal{F}_n$ and $\mathcal{W}_v^{seen}$, $\mathcal{W}_n^{seen}$ are trained via cross entropy (CE) losses $\mathcal{L}_{cls}^v$ and $\mathcal{L}_{cls}^n$ with verb~/~noun labels $y_v$, $y_n$. 

\begin{eqnarray}
\mathcal{L}_{cls}^v = \mathbf{CE}(\mathcal{F}_v; \mathcal{W}_v^{seen}; y_v) \\
\mathcal{L}_{cls}^n = \mathbf{CE}(\mathcal{F}_n; \mathcal{W}_n^{seen}; y_n)
\end{eqnarray}

We extract disentangled features for verbs and nouns, so that verbs and nouns can be treated as separate entities. If verb features contain much information about nouns, it would overfit to seen actions and would not generalize to unseen compositions. Standard networks like Inception3D (I3D) \cite{I3D} can rely on scene or object information to predict verbs~\cite{bias,somethingelse}. To decouple verb's representation from noun's, we add explicit regularization to the model input. We first used an off-the-shelf class-agnostic object detector to detect the bounding box of interacting objects. Then, we crop the object from videos and use I3D backbone to extract verb features from the cropped videos. \cite{cut_mix,cut_out,dance_in_mall,hide_and_seek} use similar cropping technique to remove bias in other tasks. We also detect hand masks and add hand regions separately to the verb input because the class-agnostic detector tends to crop out the hands as well. Adding hand gesture information back provides hints for verbs. 
Disentanglement method on Charades dataset~\cite{charades} is different as it contains third-person view videos, and relevant details are discussed in Section \ref{sec:exp:charades}.

\subsection{GCN for learning novel concept}
\label{sec:method:gcn}
After training on seen concepts, we can infer the classifier for unseen ones. In this subsection, we drop the subscript $v/n$ as the same process applies for both verb and noun. After learning feature extractor $\mathcal{F}$ and classifier of seen concept $\mathcal{W}^{seen}$, learning classifier for unseen concepts is equivalent to learning the weight $\mathcal{W}^{unseen}$. 
This step leverages graph convolution network (GCN) following previous work \cite{kampffmeyer2019rethinking,Kato_2018_ECCV,zero}. It takes the word embedding of unseen/seen concepts $s_{seen}, s_{unseen}$, and conducts graph convolution on the KG, with previously learned classifier weight $\mathcal{W}^{seen}$ as supervision. In each layer, it calculates:

\begin{equation}
 Z_{i+1} = \hat{A} Z_{i} W_i
\end{equation}

$Z_{i}$ and $Z_{i+1}$ are input and output of the layer $i$, $\hat{A}$ is the adjacency matrix of the graph. Following \cite{kampffmeyer2019rethinking,Kato_2018_ECCV,xiaolongwang}, we normalize the adjacency matrix. $W_i$ is a learnable parameter. GCN first transforms features linearly, then aggregates information between nodes via graph edges. The $0^{th}$ layer's input $Z_{0}$ is the word embedding [$s_{seen}, s_{unseen}$]. The last layer's output $Z_{n}$ is the classifier weight [$\hat{\mathcal{W}^{seen}}$, $\hat{\mathcal{W}^{unseen}}$]. We use the $\mathcal{W}^{seen}$ learned previously as supervision, and calculate the mean square error loss between $\mathcal{W}^{seen}$ and $\hat{\mathcal{W}^{seen}}$. The training process is illustrated in Figure \ref{fig:training}. Only the GCN learning of nouns is shown for brevity.

\begin{equation}
     \mathcal{L}_{GCN} = \mathcal{L}_{mse} (\hat{\mathcal{W}^{seen}}~,~\mathcal{W}^{seen})
     \label{mse}
\end{equation}

\subsection{Incorporating affordance prior}
\label{sec:method:affd}

Not all verb-noun pairs are equally important --- some objects can only admit certain actions but not others. \cite{Affordance} proposed the notion of ``affordance" --- the shape of an object may provide hints on how humans should use it, which induces the set of suitable actions. Affordance can be extracted from the language source, e.g. we will often say $peel~the~apple$ but rarely $peel~the~chair$. Prior works~\cite{Context-aware,Visual_Relationship_Detection} used language information as prior to improve their performance. In this paper, we use captions of HowTo100M dataset~\cite{howTo100} which records human-object interaction. We run the Standford NLP parser \cite{parser} to extract nouns/verbs from captions automatically. 

After extracting verb-noun pairs, we train a scoring function $\mathcal{A}$ to calculate the verb-noun affordance matching score. We project verb embedding $s_v$ to the noun embedding space and calculate cosine distance with $s_n$, followed by sigmoid to output a scalar value indicating whether this verb-noun compositional action is plausible. For training, we generate positive/negative pairs and use binary cross-entropy loss $\mathcal{L}_{affd}$. Note that there underlies an open-world assumption \cite{openworld}: the verb-noun pairs missing are not entirely infeasible, but could be unobserved. Further research can be explored to develop a more precise way of modeling the affordance constraint.

A scoring function based on only word-embedding is similar to a static look-up table for verb-noun pairs, and may fail to encode diverse action visual features. Thus we train a mapping function $\mathcal{M}$ to transform verb's visual input to its word embedding $s_v$ using mean square error loss. 

\begin{equation}
     \mathcal{L}_{mse} (\mathcal{M} (X_v), s_v)
     \label{mse_aff}
\end{equation}

\begin{figure}[t]
\centering
\includegraphics[width=0.45\textwidth]{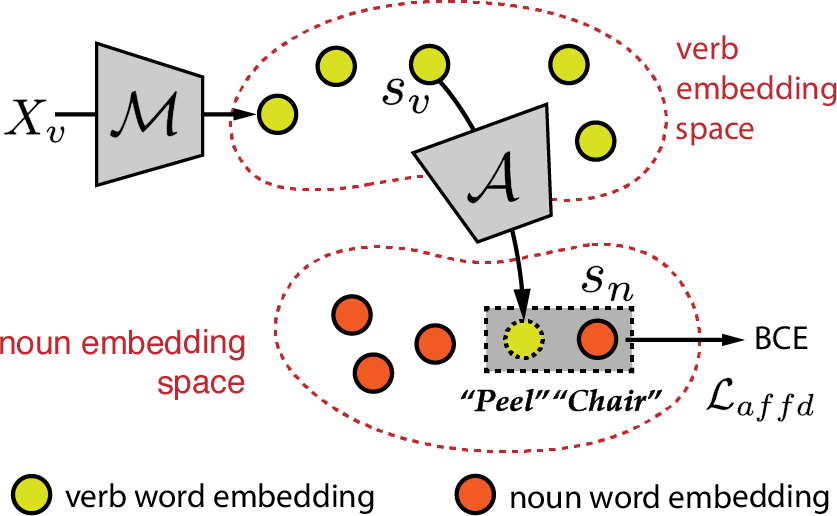}
\caption{We train a scoring function $\mathcal{A}$ to calculate the affordance matching between verb and noun. In test time, the matching score is computed between mapped verb visual feature $\mathcal{M} (X_v)$ and $s_n$.}
\label{fig:affordance}
\end{figure}

The separation of $\mathcal{A}$, $\mathcal{M}$ also adds interpretability and allows learning from different data. $\mathcal{A}$ can be trained on a language corpus without video data. Also, $\mathcal{A}$ deals with textual affordance relationship directly and adds interpretability. In test time we map verb's visual input to verb embedding space and calculate affordance score with target noun's embedding $s_n$ (Figure~\ref{fig:affordance}).
The model is asymmetric, since we use object proposals with false detection and verb visual input is more reliable.

\subsection{Overall algorithm and inference}
\label{sec:method:overall}

\begin{algorithm}[t]
\label{algo}
\smallskip
1. Train the feature extractor and corresponding classification weight ($\mathcal{F}_v$, $\mathcal{W}_v$), ($\mathcal{F}_n$, $\mathcal{W}_n$) via classification loss $\mathcal{L}_v$, $\mathcal{L}_n$ separately. \\

2. Use the word embeddings $s_v$, $s_n$ as input and the learned classification weight $\mathcal{W}_v$, $\mathcal{W}_n$ as supervision, to train the GCN model ($\mathcal{G}_v$, $\mathcal{G}_n$) with mean square error (MSE) loss (Equation \ref{mse}). \\
 
3. Use extracted affordance pairs to train scoring function $\mathcal{A} (s_v, s_n)$ (Figure~\ref{fig:affordance}), via binary cross entropy (BCE) loss $\mathcal{L}_{affd}$.\\

4. Train mapping function $\mathcal{M}$ to map visual verb inputs to semantic embedding space (Equation \ref{mse_aff}).
 
\caption{Training process of our DARK model}
\end{algorithm}

The training of our DARK model is shown in Algorithm 1. During inference, we calculate the probability of a video containing the compositional action ($v$, $n$) using following equations: 
\begin{eqnarray}
    \mathcal{P}(v,n) = \mathcal{P}(v) * \mathcal{P}(n) * \mathcal{A} (\mathcal{M} (X_v), s_n) \\
    \mathcal{P}(v) = \sigma( \mathcal{W}_v * \mathcal{F}_v (X) )
\end{eqnarray}

$\sigma$ is sigmoid function. For the classification weights $\mathcal{W}_v$ used in verb prediction $\mathcal{P}(v)$, we use the learned classification weight $\mathcal{W}_v^{seen}$ for seen classes, and $\hat{\mathcal{W}_v^{unseen}} $ predicted by GCN for unseen classes. Similar equation applies for noun prediction~$\mathcal{P}(n)$, which is omitted.

%% file: 4_experiment.tex
\section{Experiments}
\label{sec:exp}

In this section, we discuss experiment evaluation, setup and results. Some implementation details are in appendix. \footnote{Code and proposed benchmark are in  \url{https://github.com/airmachine/DARK}.}

\begin{figure}[t]
\centering
\includegraphics[width=0.4\textwidth]{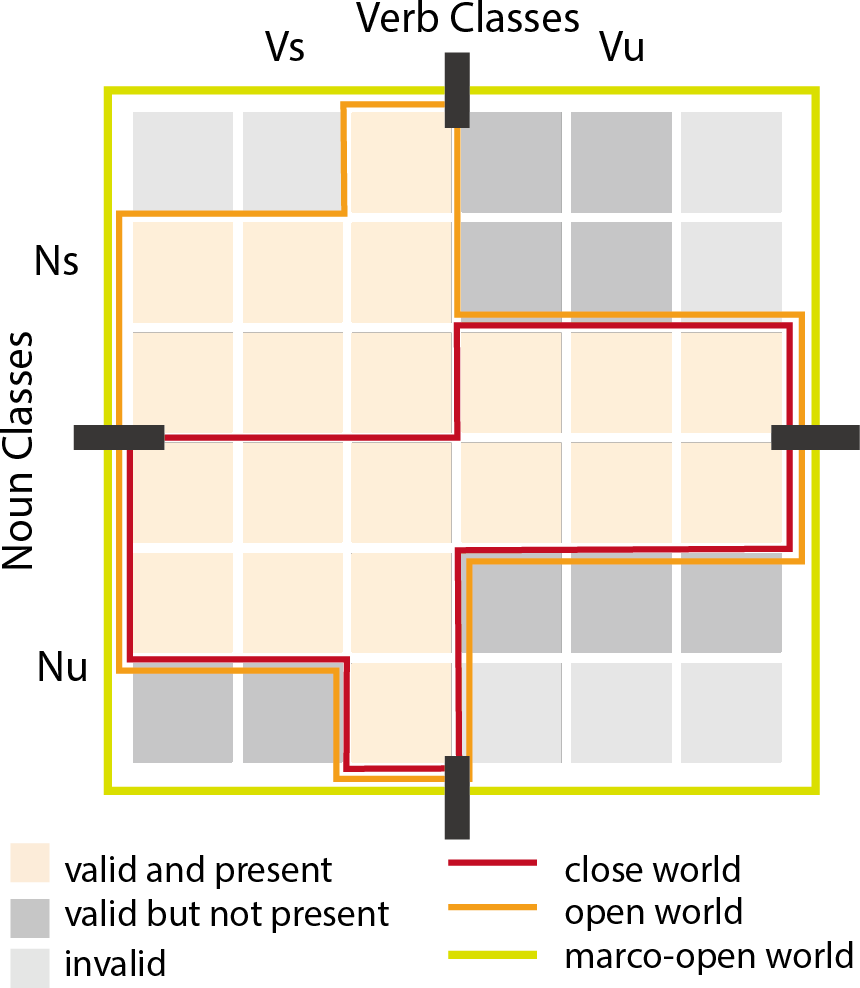}

\caption{The seen compositional actions correspond to $V_sN_s$ (the upper left part), and unseen actions include $V_sN_u, V_uN_s$ and $V_uN_u$ (the rest). We visualize the scope of close~/~open~/~macro-open world settings.}

\label{fig:eval}
\end{figure}

\subsection{Evaluation of zero-shot compositional task}
\label{sec:exp:eval}

Following previous work~\cite{Kato_2018_ECCV}, we partition verbs into two disjoint sets for seen/unseen classes, \textbf{$V_s$}~/~\textbf{$V_u$}, and same for nouns, \textbf{$N_s$}~/~\textbf{$N_u$}. Thus, ``seen compositional actions" correspond to $V_sN_s$, while ``unseen (zero-shot) actions" correspond to $V_sN_u, V_uN_s$ and $V_uN_u$.

Prior to \cite{GZSL}, most works (e.g.~\cite{conse}) on zero-shot learning adopt an evaluation protocol where predictions are made and evaluated only among unseen classes. This is later denoted as \textbf{close world} setting. \cite{GZSL} points out that it does not reflect the difficulty of recognition in practice, and there exists a trade-off between classification accuracies of seen~/~unseen classes. They propose \textbf{generalized zero-shot learning (GZL)} setting --- test set contains samples of both seen~/~unseen classes. Predictions are made and evaluated on both categories. By adding different biases to unseen classes' prediction, one can draw a curve depicting the trade-off between accuracies of seen/unseen samples. They use area under the curve (AUC) to better reflect model's overall performance on both seen and unseen classes. Our evaluation follows this setup.

Currently, relatively few prior works tackle zero-shot compositional action recognition~\cite{Kato_2018_ECCV,somethingelse}. 
Taking other zero-shot compositional learning tasks such as zero-shot attribute-object classification~\cite{red_wine,attr_operator,attribute,hierach_attr} and image-based zero-shot human-object interaction \cite{JunsongYan,HOI_knowledge} into consideration, we find that zero-shot compositional learning task poses extra challenges due to its combinatorial label space: not all compositional labels are valid, and for the valid ones, there may be no samples in the dataset. For brevity, we continue to use the term ``verb" and ``noun", but the following discussion could be also applied to other zero-shot compositional learning tasks (e.g., attribute-object).
As in Figure \ref{fig:eval}, with $N_v$ verbs, $N_n$ nouns, we can construct a $N_v \times N_n$ action label space. Among these actions, some are $invalid$, because the verb-noun pair contradicts our common sense (e.g. $peel~a~car$). And some are $valid~but~not~presented$ in the dataset. Most previous works only consider labels that contain samples in the dataset, namely compositional classes that are $valid~and~presented$ in the dataset.

\begin{table*}[!htbp]
\centering
\caption{DARK's results on Epic-Kitchen dataset, compared with baselines and GCNCL \cite{Kato_2018_ECCV} }
\vspace{-0.7em}
\scalebox{0.8}{
\begin{tabular}{c|ccc|ccc|cc}
\toprule
- &  \multicolumn{3}{c|}{AUC Macro Open} & \multicolumn{3}{c|}{AUC Open} &  \multicolumn{2}{c}{mAP Open}\\

~   &  Top1 & Top2 & Top3  &  Top1 & Top2 & Top3 & All & Zero-shot class \\

\midrule
Chance   & 6.4$\times$ $10^{-7}$ & 2.9$\times$ $10^{-6}$
        &  7.8$\times$ $10^{-6}$    
        & 1.9$\times$  $10^{-5}$  & 6.7$\times$ $10^{-5}$ 
        & 1.2$\times$  $10^{-4}$   
        & 0.00065 & 0.00067 \\
        
Triplet    & 0.021 & 0.060 & 0.095    & 0.021 &  0.060 & 0.094   & 0.051 & 0.053 \\

SES    & 0.091 & 0.25 & 0.42    & 0.16 & 0.45 & 0.73   & 1.83 & 0.99   \\

DEM    & 0.0068 & 0.019 & 0.040    & 0.022 &  0.074 & 0.14   & 0.52 & 0.30   \\

\midrule

GCNCL+GT    & 0.044 & 0.11 & 0.19    & 0.044 & 0.11 & 0.19   & 0.46 & 0.31  \\

GCNCL+Affd    & 0.064 & 0.16 &  0.27    & 0.082 & 0.22 & 0.37   & 0.48 & 0.15   \\

GCNCL+Both    & 0.061 & 0.16 & 0.26    & 0.07 & 0.17 & 0.27   & 0.47 & 0.27   \\

\midrule
DARK (ours) & 1.69 & 3.64 & 5.45    & 2.04 &  4.67 & 7.05   & 2.39 & 1.22\\
\bottomrule

\end{tabular}}
\label{table:res:epic}
\end{table*}

Prior works~\cite{somethingelse,red_wine,attr_operator,hierach_attr} use disjoint label spaces in training and test sets, which corresponds to the \textbf{close world setting}. \cite{attribute}'s test set contains both seen and unseen classes (GZL setting) and uses the AUC metric like ours, but their prediction is made and evaluated only among compositional classes with samples in dataset. \cite{Kato_2018_ECCV,JunsongYan} also follow GZL setting, but they use mean average precision (mAP) over compositional classes presented in dataset, which implicitly only considers $valid~and~presented$ classes. We denote this as \textbf{open world setting}: test set contains samples from seen/unseen classes, but prediction is made and/or evaluated among $valid~and~presented$ classes. 

Neither close world setting nor open word setting reflects the difficulty of zero-shot compositional action recognition task. When deploying recognition models in the real world, it would need to make predictions in the whole $N_v \times N_n$ label space. Thus, compositional constraints between verbs and nouns (affordance) should be properly modeled to exclude $invalid$ classes. In addition, the evaluation protocol should not distinguish between classes that are $valid~but~not~presented$ and $valid~and~presented$ in the dataset, because models would not have access to that information beforehand. We propose the \textbf{marco~open~world setting}. In test time, sample can be from all seen/unseen classes, including $V_sN_s$, $V_sN_u$, $V_uN_s$ and $V_uN_u$, and model receives no information about where the sample comes from. Predictions are made and evaluated in the whole $N_v \times N_n$ label space, and the AUC metric~\cite{GZSL} considering the trade-off between seen/unseen classes is reported. Figure~\ref{fig:eval} compares these three settings.

\subsection{Experimental setup}
\label{sec:exp:setup}

\noindent{\textbf{Dataset and split}:} 
We conduct experiments on two datasets, Epic-kitchen v-2~\cite{epic2,epic1} and Charades~\cite{charades}. On Epic-kitchen benchmark, we create the compositional split for compositional action recognition. To avoid inductive bias brought by pretrained backbones (e.g. Faster R-CNN \cite{fasterrcnn} pre-trained on ImageNet  \cite{deng2009imagenet}, or Inception3D (I3D) \cite{I3D} pre-trained on Kinetics \cite{Kinetics}) as discussed in~\cite{JunsongYan}, we ensure all nouns/verbs seen during pre-training stay in $V_sN_s$ when creating compositional split on Epic-kitchen benchmark. For Charades, we follow the same splits in~\cite{Kato_2018_ECCV} for fair comparison.

\textbf{Charades} dataset~\cite{charades} contains 9848 videos, and many involve compositional human-object interaction. We use the compositional benchmark proposed by~\cite{Kato_2018_ECCV}: they remove ``no interaction" action categories, leaving 9625 videos with 34 verbs and 37 nouns. Those verbs and nouns are further partitioned into two verb splits $V_s, V_u$ (number of classes being 20~/~14), and two noun splits $N_s, N_u$ (18~/~19). The total number of compositional actions is 149.

\begin{table}
\centering
\caption{Dataset statistics of proposed benchmark based on Epic-kitchen and previous benchmark on Charades. }
\vspace{-0.7em}
\scalebox{0.85}{
\begin{tabular}{{c|cccc|c}}
\toprule
& $V_s$$N_s$ & $V_s$$N_u$ &  $V_u$$N_s$ &  $V_u$$N_u$  &  samples \\ 

\midrule
Epic-  & 840 & 896 & 1073 & 820 & 76605  \\
Charades   & 49 & 47 & 22 & 31 & 9625  \\

\bottomrule
\end{tabular}}

\label{table:dataset}
\end{table}

\textbf{Epic-kitchen version 2} dataset~\cite{epic2,epic1} contains videos recorded in kitchens, where people demonstrate their interaction with objects like pan, etc. The diversity of actions in this dataset makes it especially challenging. We follow the steps in similar previous works~\cite{zero-split,JunsongYan} to create our compositional split. We first make sure that classes seen in pre-training stay in the seen split. Then for the remaining classes we sort them based on the number of instances in descending order, and pick the last 20\% to be unseen classes, because \cite{zero-split} pointed out that zero-shot learning targets the classes not easy to collect (especially those in the tail part of the long tail class distribution). We show dataset statistics in Table~\ref{table:dataset}. We get a total number of 76605 videos, including 90 verbs, 249 nouns, and 3629 compositional actions. Compared to Charades, our proposed benchmark is at a larger scale in terms of classes involved and sample size. 

\smallskip

\noindent{\textbf{Baselines}:} 
We establish our baselines following previous work~\cite{Kato_2018_ECCV}.
Here we briefly summarize their architectures, and readers can refer to~\cite{Kato_2018_ECCV} or original papers for details. These baselines are based on Inception3D features.

\textbf{Triplet Siamese Network (Triplet)} by \cite{Kato_2018_ECCV}: verb/noun embeddings are concatenated, and transformed by fully connected (FC) layers. The output is concatenated with visual features to predict scores through one FC layer with the training of BCE loss. 

\textbf{Semantic Embedding Space (SES)} \cite{SES}: The model projects visual features into embedding space through FC layers and then matches output with corresponding action embeddings (average of verb/noun embeddings) using L2 loss.

\textbf{Deep Embedding Model (DEM)} \cite{DEM}: Verb/noun embeddings are transformed separately via FC layers and summed together. Then output is matched with visual features via L2 loss.

\subsection{Results on Epic-kitchen dataset}
\label{sec:exp:epic}

The results of the proposed DARK model, as well as the aforementioned baselines (Triplet, SES and DEM) and previous model GCNCL~\cite{Kato_2018_ECCV} on the Epic-kitchen dataset are listed in Table~\ref{table:res:epic}. We report the results in the proposed AUC metric~\cite{GZSL} with precision calculated at top 1/2/3 prediction for both open world and macro open world settings, which evaluates the overall trade-off between seen/unseen class. We also report the mean average precision (mAP) used in \cite{Kato_2018_ECCV} on all and zero-shot compositional action classes for reference.

Our best performing DARK model outperforms all baselines and GCNCL by a large margin under all metrics, illustrating the benefit of disentangled action representation for compositional action recognition. DARK is also more scalable, and reduces the number of graph nodes from 22749 (GCNCL with no external knowledge) to 339 (ours).

DARK considers the type constraint of verbs and nouns when composing verb and noun into compositional action label by training an affordance scoring module, while GCNCL considers the constraint when building compositional action nodes by collecting the existing verb-noun pairs from NEIL~\cite{NEIL}. For fair comparison, we re-implement three versions of KG in GCNCL model. In ``GT", we use the ground-truth verb-noun relationships that are presented in the dataset (open world setting). In ``Affd", we only consider relationships in the same corpus with DARK. We use the relationships as a hard look-up since GCNCL only contains unweighted ``hard" edges in its knowledge graph. In ``Both", we use the union of the constraints in ``GT" and ``Affd". Under all the three circumstances, our DARK model outperforms other models by a large margin. For all experiments, we report the best results.

\subsection{Ablations on different components}
\label{sec:exp:ablation}

\noindent \textbf{Zero-shot learning in verb/noun classifier}: 
In DARK model, we do separate verb and noun classification in two branches. We investigate different implementations of zero-shot learning in verb/noun classifier.
Specifically, we consider three options for both verb and noun, namely ``KG", ``SES" and ``ConSE".
``KG" stands for zero-shot learning by using knowledge graph to predict classification weights for unseen component with GloVe embedding \cite{GloVe} as in~\cite{Kato_2018_ECCV}. ``SES" \cite{SES} is the best common embedding baseline in Table~\ref{table:res:epic} using better BERT word embedding \cite{BERT} (based on observation in Table~\ref{table:res:affd}, BERT tends to have better performance). ``ConSE"~\cite{conse} learns a semantic structure aware embedding space compared to original word embeddings, which is modeled with graph. ``ConSE"~\cite{conse} is used as the zero-shot learning component in previous image-based action recognition task~\cite{HOI_knowledge}. It learns a semantic structure aware embedding space and we also use GloVe embedding. 
For better comparison of zero-shot learning component, we report the “open world” AUC on the Epic-kitchen dataset without using affordance (same as ``ground-truth" affordance in macro open setting), thus excluding the influence of affordance prior. Different zero-shot learning combinations for verbs and nouns are reported in Table~\ref{table:res:source}. Using ``KG" for both verb~/~noun outperforms others by a large margin, and we take this approach in the rest experiments.

\begin{table}
\centering
\caption{Combination for zero-shot learning in verb and noun classifiers. The top1 ``open world" AUC is reported on the Epic-kitchen dataset.}
\vspace{-0.7em}
\scalebox{0.85}{

\begin{tabular}{c |ccc}
\toprule
~   &  Verb-KG & Verb-SES   & Verb-ConSE  \\
\midrule
Noun-KG   &  1.81 & 1.24  &  0.49  \\
Noun-SES   &  0.40 &  0.44   &  0.32  \\
Noun-ConSE  &  0.25  & 0.18   & 0.13    \\
\bottomrule
\end{tabular}}
\label{table:res:source}
\end{table}

\noindent \textbf{Construction of verb knowledge graph}: 
Compared to nouns, the concept of verb is relatively abstract and the relationship between verbs is hard to capture. We explore different ways of constructing the verb KG, namely, ``WN dis", ``VN group" and ``VN tree". (The details of noun KG are discussed in the appendix.) In ``WN dis", we use WordNet \cite{WordNet} structure and add edges between nodes if their LCH \cite{LCH} distance is bigger than a threshold. We also explore VerbNet \cite{verbnet} which is designed to capture the semantic similarity of verbs. VerbNet categorizes verbs into different classes, and each class contains multiple verb members. To resolve the duplication in each class, we add edges between verbs in the same class, and denote this as ``VN group". We also try adding a meta node for each class and connecting all its members to the meta node, denoted as ``VN tree". Graphs of ``WN dis" and ``VN group" are naturally undirected. For ``VN tree", we consider an additional ``two-way" setting as in~\cite{kampffmeyer2019rethinking}, where a GCN model separates the parent-to-children and children-to-parent knowledge propagation into two stages to better model hierarchical graph structure. However, we do not observe performance improvement in this setting. In Table~\ref{table:res:verb}, we report top1 AUC for verbs using different KGs under the GZL~\cite{GZSL} setting. ``VN tree" in ``one-way" gives the best prediction for verb, and we keep this configuration in rest experiments.

\begin{table}
\centering
\caption{Different verb knowledge graphs. We report the top1 AUC for verbs under GZL setting.}
\vspace{-0.7em}
\begin{tabular}{{c|cccc}}
\toprule
~   &  WN dis & VN group   & VN tree \\
\midrule
one-way   &  1.86 & 0.83    & 1.93\\
two-way   &  ×  & ×   & 1.79\\
\bottomrule
\end{tabular}
\label{table:res:verb}
\end{table}

\noindent \textbf{Affordance learning}: 
We consider the compositional constraints between verbs and nouns (affordance) when composing the compositional action. We explore various ways of learning affordance in Table \ref{table:res:affd}. In ``Word-only" we train a word-embedding only model. And ``Visual" represents the approach in method section where an additional projection module maps visual features to embedding space. 
For each, we explore two word embeddings, GLoVe \cite{GloVe} and BERT \cite{BERT}. In terms of score calculation, we try three different methods each. In ``Concat-Scoring", we concatenate verb/noun features, and train a scoring model. In ``Context-Scoring", instead of concatenating, for BERT the scoring model embeds verb-noun phrase together and averages their embeddings, and for GloVe we simply average their embeddings. In ``Proj-Cosine", we project verb embedding to noun embedding space and calculate the cosine distance. We also try a lookup table, where affordance is one if the compositional pair exists in train set or knowledge bases, and zero otherwise. ``Uniform" sets all affordance to be one, 
which means no weighting is applied. ``Ground Truth" sets one for pairs existing in the dataset (train/test), equivalent to ``open world". In all experiments, we use best configuration from Table \ref{table:res:source}, and label compositional pairs seen in training to one. BERT constantly improves affordance relationships in different methods. Lookup table performs worse than ``Uniform" (no affordance) since some valid pairings are missing in knowledge bases.

\begin{table}[t]
\centering
\caption{Top1 ``marco open world" AUC on Epic-kitchen based on different affordance learning methods.}
\vspace{-0.7em}
\scalebox{0.85}{
\begin{tabular}{{c|cc|cc}}
\toprule
 &  \multicolumn{2}{c|}{Word-only} & \multicolumn{2}{c}{Visual}\\
~   &  GLoVe & BERT   & GLoVe  & BERT \\

\midrule
Concat-Scoring   &  1.49 & 1.59   & 1.48 & 1.60\\
Context-Scoring   &  1.49 & 1.59  & 1.49  & ×\\
Proj-Cosine   &  1.42  &  1.59   & 1.42  & 1.69 \\

\midrule
Lookup Table &  \multicolumn{4}{c}{1.35} \\

\midrule
Uniform &  \multicolumn{4}{c}{1.46} \\

\midrule
Ground Truth &  \multicolumn{4}{c}{1.81} \\
\bottomrule
\end{tabular}}
\label{table:res:affd}
\end{table}

\begin{table}[t]
\centering
\caption{Results (mAP) on Charades under GZL setting. Baselines and GCNCL results from \cite{Kato_2018_ECCV}.}
\vspace{-0.7em}
\begin{tabular}{{c|c|cc}}
\toprule

~   &  Model & All   & Zero-shot   \\

\midrule
~   &  Chance & 1.37  & 1.00 \\
Baseline   &  Triplet & 10.41  & 7.82 \\
  &  SES  &  10.14   & 7.81   \\
~   &  DEM  &  9.57   & 7.74  \\

\midrule
GCNCL   &  GCNCL-I+A  &  10.48   & 7.95   \\
   &  GCNCL+A  & 10.53   & 8.09   \\
   
\midrule
 Ours   &  DARK  &  11.21   & 8.38   \\
 
\bottomrule
\end{tabular}
\label{table:res:charades}

\end{table}

\begin{figure}
\centering
\includegraphics[width=0.48\textwidth]{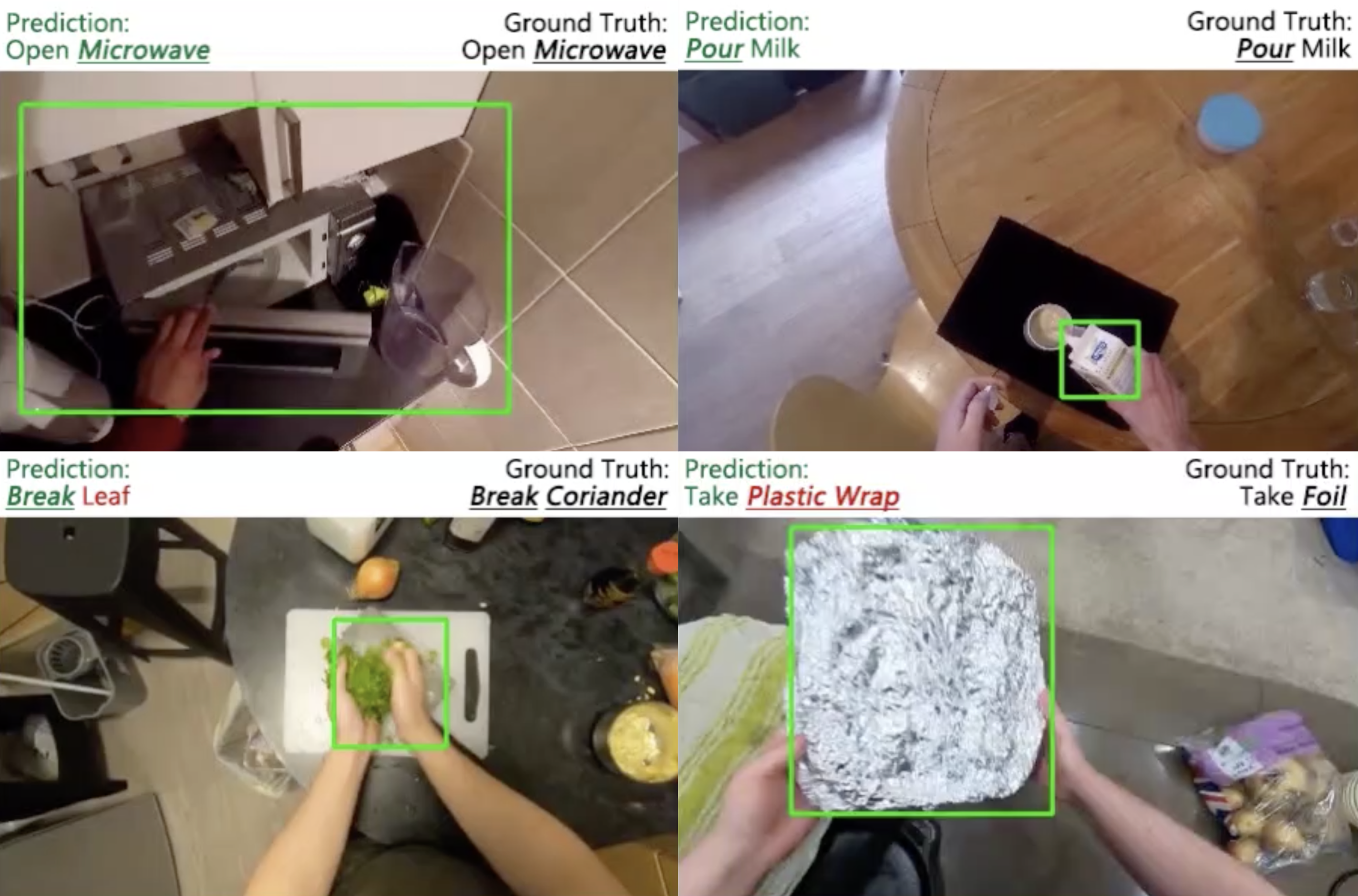} 

\caption{Qualitative Analysis. Underlines are unseen concepts, green for right predictions and red for wrong.}
\label{fig:error}
\end{figure}

\subsection{Results on Charades dataset}
\label{sec:exp:charades}
We report results on Charades in Table \ref{table:res:charades}. Following \cite{Kato_2018_ECCV}, we report mean average precision (mAP) and compare our model to theirs and baselines. We also report for zero-shot classes ($V_sN_u+V_uN_s+V_uN_u$) separately but all predictions are made under GZL setting. Unlike Epic-Kitchen which contains ego-centric actions, Charades contains third-person view videos and cannot detect the mask of person's hand. Thus we directly learn the verb and noun feature disentanglement leveraging a discriminator and a disentanglement loss. Following \cite{discriminator}, discriminator tries to adversarially classify noun label $y_n$ from its verb feature, and feature extractor $\mathcal{F}$ goes against it. To better capture the multi-label property in Charades, we use an un-factorized classification model for actions in $V_sN_s$ so they can be treated separately. Since we report the mAP results for fair comparison with GCNCL, we do not use affordance in our model. As indicated in~\cite{Kato_2018_ECCV}, we also notice that the amount of improvement over baselines is not large, possibly because Charades is relatively small and easy to overfit. And this motivates us to propose a large-scale zero-shot compositional action recognition benchmark.

\subsection{Qualitative error analysis}
\label{sec:exp:error}
\noindent We also visualize some examples in Figure \ref{fig:error}. The model misclassifies \textit{coriander} as \textit{leaf}, and \textit{foil} as \textit{plastic wrap} due to visual similarity.

%% file: 5_conclusion.tex
\section{Conclusion}
\label{sec:conclusion}
In this paper, we propose DARK, a novel compositional action recognition model that reduces complexity from quadratic to linear, making the training more scalable. DARK generalizes to unseen verb-noun pairs, and can be combined with knowledge bases to produce state-of-the-art compositional action recognition results.

%% file: supp_1_implementation.tex
\section{Implementation Detail}

To build the object feature extractor $\mathcal{F}_n$, we first use an off-the-shelf class-agnostic object detector to detect the bounding box of interacting objects. For the Charades \cite{charades} dataset, we use the detected object boxes generated by HOID~\cite{JunsongYan}. HOID model first detects the box of human, then based on the human's bounding box it detects interacting objects. We use the code publicly released by \cite{JunsongYan}, with the default parameters provided by the author. We use the weights of the class-agnostic object detector provided on the project page. For the Epic-kitchen dataset \cite{epic2,epic1}, we use the pre-computed class-agnostic boxes provided by the authors. They use the model from \cite{Shan2020UnderstandingHH}, which detects human hands and locates the interacting objects. We use different models for object detection because Epic-kitchen \cite{epic1} contains mostly ego-centric action videos while Charades \cite{charades} contains third-person view videos. For the Epic-kitchen dataset, we additionally use its pre-computed detected masks of human hands. 

After the object boxes are detected, we run the Faster-RCNN \cite{fasterrcnn} model with the ResNet-101 backbone to extract the object features. We use the implementation provided by the Detectron2 \cite{wu2019detectron2} library, with the weights pre-trained on ImageNet \cite{deng2009imagenet}. We obtain the features before the classifier layer in the Faster-RCNN model, which results in 2048 dimension object features. If we detect multiple boxes in one frame, we conduct max-pooling over their extracted features to obtain one feature representation for each frame. If no box is detected for a particular frame, we then extract the feature of the whole image as its object feature.

We sample several frames along the temporal axis of the video to conduct object detection. Since videos in the Charades \cite{charades} dataset may contain more than one action, we treat each frame in Charades dataset as one sample for training. For the Epic-kitchen \cite{epic2,epic1} dataset whose videos contain only one action, we instead simply apply mean pooling over object features of sampled frames to obtain one feature representation for the whole video. We add fully-connected (FC) layers upon the fixed Faster-RCNN backbone to conduct feature extraction.

For verb features, we build our feature extractor $\mathcal{F}_v$ based on a standard two-stream Inception3D~\cite{I3D} backbone pre-trained on the Kinetics dataset~\cite{Kinetics}. We use both the RGB branch and the optical-flow branch, each producing a 1024 dimension feature in the layer Mixed\_5c. We then concatenate them, resulting in a feature representation of 2048 dimension. For the Epic-kitchen dataset, we generate features first using the video input with the object cropped out. Then we do the same using the video input with everything cropped except for the detected hands in order to obtain hand gesture movement information. We further concatenate these two features to get a 4096 dimension feature. Similar to the object feature extractor, we add FC layers to features generated by the fixed Inception3D backbone. For the Charades dataset, since another disentanglement approach is used, we simply use the 2048 dimension feature.

Following common practice, we split the whole video into video clips with a small duration, and generate features for each clip during training and inference. For the Charades dataset, we sample 10 clips per video to conduct training and we treat each clip as a sample. Whereas for the Epic-kitchen dataset, we apply max pooling to the features of all the clips generated from one video to obtain one feature representation for the whole video.

Our model is implemented in PyTorch with Adam optimizer. We used in total around 20 GPUs through out the experiments. But a single run only needs 5 GPUs. (we launch parallel experiments)

%% file: supp_2_epic.tex
\section{The Proposed Epic-kitchen Benchmark}

We build our compositional action recognition benchmark based on the Epic-kitchen~\cite{epic2,epic1} dataset version two. We take the class that overlaps with pre-trained backbones into consideration when creating seen/unseen class splits. We find that there are 95 noun classes overlapping with ImageNet classes, and 23 verb classes overlapping with Kinetics classes, where the backbones that we use have been pre-trained. We make sure these overlapping classes stay in the seen split.

We then remove the tail verb and noun classes with less than 10 instances. The remaining dataset contains a total number of 76605 videos, including 90 verbs, 249 nouns, and 3629 compositional actions. We have 29 verbs in the seen category, and 61 verbs in the unseen category. On the other hand, 102 nouns are seen and 147 nouns are unseen.    

The $V_sN_s$ split contains 840 compositional actions, and 51228 samples. The $V_uN_s$ split has 1073 compositional actions, and 10105 samples. For the $V_sN_u$ split, there are 896 compositional actions and 11073 samples. And for the $V_uN_u$ split, there are 820 compositional actions and 4199 samples.

Epic-Kitchen dataset is built under the Creative Commons Attribution-NonCommercial 4.0 International License. The licence for non-Commercial use of Charades dataset can be found at http://vuchallenge.org/charades.html. We follow the intended usage of these two datasets.

%% file: supp_3_noun_KG.tex
\section{Noun Knowledge Graph Construction}
\label{sec:dataset}

We discuss the construction of the verb knowledge graph in the paper, due to the space limit, we present the details of the noun knowledge graph in this section. We construct the noun knowledge graph following \cite{xiaolongwang}'s approach. We begin from the nouns presented in the dataset, and recursively search their hyper-norms using WordNet \cite{WordNet}'s lexical relationship to add to the graph. In addition, we augment the knowledge graph by adding nouns from ImageNet's \cite{deng2009imagenet} class labels.

When building noun knowledge graphs, we add an edge if two entities are direct synonyms or hyper-norms. Our model is built upon the graph convolution model implemented by \cite{kampffmeyer2019rethinking}. We use its plain GCN version without attention. And we use the ``two-way" approach, which separates parent-to-children and children-to-parent knowledge propagation into two stages to better model the hierarchical graph structure. For noun knowledge graph learning, we use 300d GloVe \cite{GloVe} embeddings as input.

%% file: supp_4_disentanglement.tex
\section{Disentanglement in Charades Dataset}
\label{sec:dataset}

Let $X_v$ denote the input to the verb feature extractor $\mathcal{F}_v$, and $X_v'$ denote the extracted verb features. Similarly, $X_n$ is the input to $\mathcal{F}_n$ and $X_n'$ is the extracted noun features.

\begin{eqnarray}
X_v' = \mathcal{F}_v (X_v) \\
X_n' = \mathcal{F}_n (X_n) 
\end{eqnarray}

To obtain disentangled verb~/~noun features, we take the idea from the previous paper \cite{discriminator}. We use a discriminator to limit the information verb and noun features contain. The discriminator $\mathcal{D}_v$ tries to adversarially classify noun label $y_n$ from its verb feature $X_v'$, and the feature extractor $\mathcal{F}_v$ goes against it via a minimax process. The discriminator helps to limit the information which verb feature $X_v'$ contains about the nouns in the video. The same procedure happens for $\mathcal{D}_n$. We use one layer linear classifier for discriminator $\mathcal{D}_{v}$ and $\mathcal{D}_{n}$, and they output class predictions for the opposite branch. This leads to the disentanglement loss:

\begin{eqnarray}
\mathcal{L}_{dis}^{v} = -\mathbf{CE}(\mathcal{D}_{v}(X_v'); y_n) \\
\mathcal{L}_{dis}^{n} = - \mathbf{CE}(\mathcal{D}_{n}(X_n');y_v)
\end{eqnarray}

The $\mathbf{CE}$ refers to the cross-entropy loss. The overall loss for training the feature extractor $\mathcal{F}_v$, $\mathcal{F}_n$ and the classifier for seen classes $\mathcal{W}_v^{seen}$, $\mathcal{W}_n^{seen}$ is:

\begin{eqnarray}
     \mathcal{L}_v =~ \mathcal{L}_{cls}^v  + \mathcal{L}_{dis}^{v}   \\
     \mathcal{L}_n =  \mathcal{L}_{cls}^n  + \mathcal{L}_{dis}^{n}
\end{eqnarray}

The definitions of $\mathcal{L}_{cls}^v$, $\mathcal{L}_{cls}^n$ are the same as discussed in the main paper.